\pdfoutput=1

\documentclass[11pt]{article}

\usepackage[final]{acl}

\usepackage[utf8]{inputenc}
\usepackage[T1]{fontenc}
\usepackage{times}
\usepackage{microtype}
\usepackage{latexsym}
\usepackage{svg}

\usepackage{hyperref}       
\usepackage{url}            
\usepackage{booktabs}       
\usepackage{nicefrac}       
\usepackage{fancyhdr}       
\usepackage{graphicx}       
\usepackage{enumitem}       
\usepackage{amsfonts}       
\usepackage{amsmath}        
\usepackage{amssymb}        
\usepackage{bm}             

\graphicspath{{img/}}

\title{Uniform Masking Prevails in Vision-Language Pretraining}
  
\author{
Siddharth Verma$^1$, Yuchen Lu$^2$, Rui Hou$^1$, Hanchao Yu$^1$, Nicolas Ballas$^1$,\\
\bf Madian Khabsa$^1$, Amjad Almahairi$^1$\\
$^1$Meta, AI Integrity, $^2$University of Montreal, Mila \\
}

\begin{document}
\maketitle

\begin{abstract}
Masked Language Modeling (MLM) has proven to be an essential component of Vision-Language (VL) pretraining. To implement MLM, the researcher must make two design choices: the masking strategy, which determines which tokens to mask, and the masking rate, which determines how many tokens to mask. Previous work has focused primarily on the masking strategy while setting the masking rate at a default of 15\%. In this paper, we show that increasing this masking rate improves downstream performance while simultaneously reducing performance gap among different masking strategies, rendering the uniform masking strategy competitive to other more complex ones. Surprisingly, we also discover that increasing the masking rate leads to gains in Image-Text Matching (ITM) tasks, suggesting that the role of MLM goes beyond language modeling in VL pretraining.
\end{abstract}

\section{Introduction}
Recently, large-scale multi-modal transformers~\cite{vilt,meter,albef,uniter,tan2019lxmert,vilbert} have dominated the field of Vision-Language (VL) pretraining. These models are usually trained via a multi-task objective, with the most prominent subtask being Masked Language Modeling (MLM). MLM is a self-supervised learning objective that was made popular by \citet{devlin2018bert}. Within MLM, the input tokens are corrupted and fed to the model, which learns to reconstruct the input. Crucial to implementing MLM is the \textit{masking strategy}, which selects a subset of tokens to be corrupted. Different masking strategies are characterized by which parts of the input they choose to mask. Most common strategies are further parameterized by the \textit{masking rate}, which determines what percentage of the input to mask. 

\begin{figure*}
    \includegraphics[width=\linewidth]{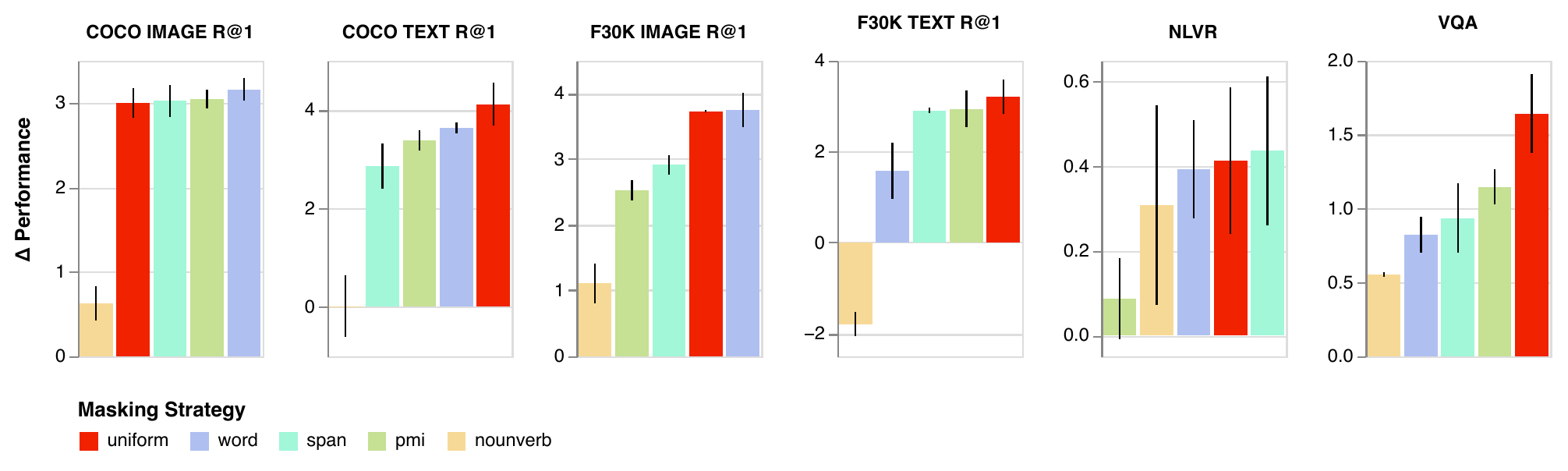}
    \caption{Change in performance when increasing masking rate from 15\% to 60\% across various tasks and masking strategies. Error bars show standard error of the mean. All results but one are positive, indicating that increasing the masking rate improves performance. Furthermore, uniform masking (shown in red) attains highest or close-to-highest performance increases across all tasks, highlighting its scalability.}
    \label{img:results_delta}
\end{figure*}

Prior studies indicated that the masking strategy played a vital role in improving downstream performance. For example, masking named-entities was shown to improve natural language inference~\cite{sun2019ernie}, and PMI masking~\cite{levine2020pmi} claimed better performance than uniform masking, the default masking strategy of BERT~\cite{devlin2018bert}. Naturally, when MLM was introduced to VL pretraining, researchers explored these strategies alongside VL-specific ones like noun-verb masking~\cite{bitton2021data}, and found that the masking strategy had a noticeable impact on performance.

Although the masking strategy seems to have an important role on downstream performance, adjusting the task difficulty can achieve similar effects. For example, ELECTRA~\cite{clark2019electra} demonstrated that changing the prediction problem can increase performance. \citet{wettig2022should} showed that increasing the masking rate for text-based language modeling leads to better downstream performance. We hypothesize that increasing the masking rate for vision-language pretraining is a simple and effective way to create a harder learning problem, thereby increasing downstream performance and reducing performance differences across masking strategies.

In this paper, we investigate the effect of the masking rate on VL pretraining across various masking strategies. Our contributions are threefold:

\begin{enumerate}
    \item We show that increasing the masking rate leads to performance gains on VL tasks across various masking strategies.
    \item We empirically find that uniform masking performs competitively with other complex masking strategies at high masking rates, indicating its scalability with respect to masking rate.
    \item We show that increasing the masking rate leads to better image-text retrieval performance, suggesting that the role of MLM goes beyond purely caption modeling.
\end{enumerate}

\section{Preliminaries}
\subsection{Vision-Language Pretraining}
Vision-Language Pretraining is generally accomplished via a multi-task objective consisting primarily of three types of losses: language, vision, and alignment loss. For language losses, most implementations use either casual language modelling~\cite{radford2019language} or masked language modelling~\cite{devlin2018bert}. The alignment loss aims to leverage the pairing information of images and captions. It is usually implemented in one of two methods: as an Image-Text Matching loss over a binary classifier~\cite{vilt, albef}, or as a contrastive loss~\cite{radford2021learning}. Finally, recent work~\cite{singh2022flava} investigates the use of the masked autoencoder loss~\cite{kaiming2021mae} as a vision loss.

\subsection{Masked Language Modeling}
MLM is an unsupervised objective used to train language models that was made popular by BERT~\cite{devlin2018bert}. Within MLM, a subset of the input tokens are corrupted and the model is trained to predict the original tokens. This subset is chosen by a masking strategy. In BERT, 15\% of the input tokens are randomly chosen to be corrupted. Finally, the input is corrupted by a corruption strategy. In BERT, each token has an 80\% chance of being replaced with the {\tt [MASK]} token, 10\% chance of being replaced with a random word and 10\% chance of being unchanged.

\label{sec:masking_method}
\subsection{Masking Strategies}
Let a mask $M$ be the vector $[m_1,m_2,\cdots,m_n]$ where $m_i$ is either $0$ or $1$. Furthermore, let the input $T$ be a vector of tokens $[t_1,t_2,\cdots,t_n]$ where $t_i\in\mathcal T$ and $\mathcal T$ is the set of all tokens. Then, a masking strategy $f$ is defined as the conditional probability $P(M|T)$. In this paper, all masking strategies considered have a tunable parameter $p$ which controls the masking rate, such that \(\mathbb E [\hat p] \approx p \), where \( \hat p = \frac 1 n \sum_{i=1}^n m_i \) is the empirical masking probability.

\paragraph{Uniform Masking Strategy}
This is the simplest masking strategy where each token is masked independently with probability $p$. Thus, \( f = [{\rm Bernoulli}(p)]_n \).

\paragraph{Whole Word Masking Strategy}
This is similar to uniform masking, except that a unit is a word rather than individual tokens. Specifically, given a mapping $w:\{1..n\}\to\{1..m\}$ which maps the $i$th token to the $w(i)$th word, construct the matrix $W\in\mathbb{R}^{n\times m}$ such that $W_{ij}={\bf1}\{w(i)=j\}$. Then, \( f = W[{\rm Bernoulli}(p)]_m \).

\paragraph{Noun-Verb Masking Strategy}
Originally proposed by \citet{bitton2021data}, Noun-verb masking in its original form only masks the nouns in a sentence. As this strategy cannot reach a high empirical masking rate, we modify it to mask tokens in order of categories until we have masked $np$ tokens. We set the order as follows: nouns, proper nouns, verbs, adjectives, adverbs and all other tokens.

\paragraph{Span-based Masking Strategy}
Originally proposed by \citet{joshi2020spanbert}, span-based masking masks spans of text rather than dispersed tokens. Spans are chosen with a random starting point and length $X|X\le{}10$ where $X\sim{\rm Geom}(0.2)$. They are repeatedly chosen and masked until the empirical masking rate $\hat p \ge p$.

\paragraph{PMI-based Masking Strategy}
PMI masking, introduced by \citet{levine2020pmi}, masks ngrams with a higher Pointwise Mutual Information (PMI) over ngrams with a lower PMI. PMI is defined as
\[
{\rm PMI}(x_1,x_2,\ldots,x_n) 
= \frac
  {p(x_1,x_2,\ldots,x_n)}
  {\min_{s\in\sigma(x)}\prod_{\hat x\in s}p(\hat x)},
\]
where $\sigma(x)$ is all the partitions of $x$ not including $x$ itself. To implement PMI masking, we first we compute the PMI for all ngrams within the training set where $n\in\{2, 3, 4, 5\}$. Then, for each $n$ we rank the ngrams in descending order of their PMI and select the top $800,000$. To mask a sentence, we group the input tokens according to the computed ngrams and mask each group with probability $p$.
\section{Experiment Setup}
\subsection{ViLT Architecture}
In this work, we perform analysis using ViLT~\cite{vilt}, a recent transformer-based VL model that achieves superior performance on a variety of downstream tasks. We chose ViLT because of its strong performance, simple architecture, and open-source\footnote{\url{https://github.com/dandelin/ViLT}} implementation.

ViLT is comprised of a large fusion transformer and two small modality encoders. The fusion transformer follows the architecture of ViT~\cite{dosovitskiy2020vit} and is initialized from the \texttt{vit-base} configuration, which equates to 135M parameters. The text encoder is an embedding layer and the image encoder is a image patch layer in the style of ViT. Each input token is a summation of 3 embeddings: the modality encoder embedding, a positional embedding and a modality-type embedding indicating whether the input is a text or image. Each modality is then prepended with a \texttt{[CLS]} token, concatenated, and fed through the fusion layer to produce joint embeddings.

\subsection{Pretraining}
For pretraining, we use four datasets: Visual Genome~\cite{krishna2016vg}, COCO~\cite{lin2014coco}, SBU~\cite{ordonez2011sbu} and Conceptual Captions~\cite{sharma2018gcc}, which accumulate to a size of 4 million image-text pairs. Each pretraining run utilized 4 nodes where each node contains 96 CPU cores and 8 Nvidia A100 GPUs with 32GB of memory. The effective batch size is 4096, with a batch size of 64 per GPU. It takes 2 days and 10 hours to finish pretraining. We follow pretraining methodology from ViLT~\cite{vilt}. All hyperparameters are listed in the appendix in table \ref{tbl:hyperparams}.

\subsection{Finetuning tasks}\label{sec:finetuning_tasks}
To evaluate the model, we consider four downstream tasks listed below.

\begin{enumerate}
\item {\bf VQA2}~\cite{goyal2017vqa}: The Visual Question and Answering dataset consists of textual questions based on a provided image. Downstream score is accuracy on the dev set.
\item {\bf NLVR2}~\cite{suhr2019corpus}: The Natural Language for Visual Reasoning dataset consists of triplets of one sentence and two images $(s, i_1, i_2)$. The task is to determine whether the sentence is true given the images. To predict the answer, we concatenate representations of $(s,i_1)$ and $(s,i_2)$ and train a binary classifier using these embeddings. Downstream score is accuracy on the test set.
\item {\bf Flickr30k}~\cite{young2014flickr}: The Flickr30k dataset consists of image and caption pairs. The task is to retrieve the correct image given a caption and vice-versa. ViLT solves this by using the ITM head to predict how aligned an image and caption are. Downstream score is top-$k$ retrieval on the test set for both images and text, where we set $k\in\{1,5,10\}$.
\item {\bf COCO}~\cite{sharma2018gcc}: COCO, short for Common Objects in Context, is a dataset of image and caption pairs. The task is to retrieve the correct image given a caption and vice-versa. ViLT solves this by using the ITM head in a similar fashion to Flickr30k. Downstream score is top $k$ retrieval for both images and text, where $k\in\{1,5,10\}$
\end{enumerate}

Each finetuning run utilized a single node with the same configuration as above. The data was augmented with RandAugment~\cite{randaug} without color inversion and cutout as described in the ViLT paper. The per-gpu batch size varied per task as shown in figure \ref{tab:finetune_batch_size}. It takes a maximum of 4 hours to finish finetuning.

\label{tab:finetune_batch_size}
\begin{table}
    \centering
    \begin{tabular}{lrr}
        \hline 
        \bf Task & \bf Per-GPU & \bf Effective \\
        \hline
        VQAv2 & 128 & 1024 \\
        Flickr30k & 8   & 1024 \\
        NLVR2 & 32  & 256 \\
        \hline
    \end{tabular}
    \caption{Finetuning batch sizes for different tasks}
    \label{tbl:finetuning}
\end{table}

\section{Experiments}
Our main experiment is a grid sweep across masking rates and masking strategies. The masking rate is tuned across $[0.15, 0.3, 0.45, 0.6, 0.75]$, while the masking strategy is tuned across those listed in section $\S$\ref{sec:masking_method}. For each configuration, we pretrain ViLT and finetune it on each downstream tasks with three different seeds. We list the results for $k=1$ retrieval since the results for $k=5$ and $k=10$ are very similar. The additional results can be found in the appendix.
\subsection{Increasing the masking rate improves performance}
\begin{figure*}
\centering
  \includegraphics[width=\textwidth]{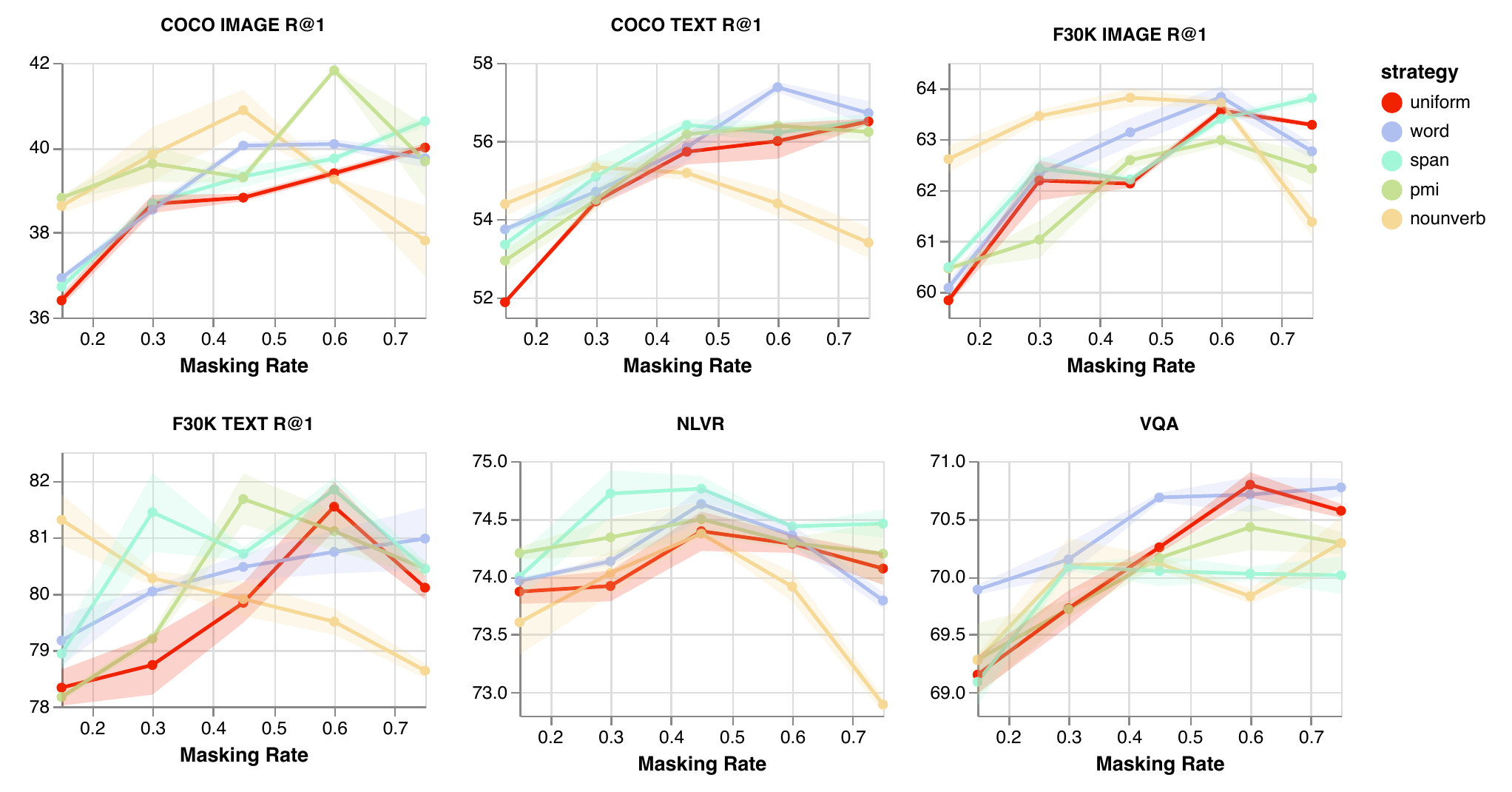}
  \caption{Downstream tasks performance vs masking probability across all strategies. Bars show standard error of the mean. Trend is generally upwards, indicating that increasing masking rate leads to performance gains on a suite of downstream tasks.}
  \label{img:results_all}
\end{figure*}

To understand how the masking rate affects performance, we study the results of the grid sweep in figure \ref{img:results_all}. A cursory glance at the figure reveals that as we increase the masking rate, performance increases until a threshold, after which it drops. Moreover, This threshold appears to vary per task; for VQA it is 75\%, for NLVR it is 45\%, for Flickr image and text retrieval it is 60\% and for COCO image and text retrieval it is 75\%. In all cases, this threshold is well above the standard 15\% recommended by BERT, demonstrating that increasing the masking rate improves downstream performance across all masking strategies and tasks.

\citet{wettig2022should} find that a masking rate of 40\% is optimal for text-based pretraining. However, our experiments suggest that a higher masking rate is needed for VL pretraining. We hypothesize that this is because of the additional information contained in the image. VL pretraining involves joint training across both images and text. Since information is present in both the text and image and we only mask the text, a higher masking rate is required to obtain a similar information density to the text-only regime.

\subsection{Quantifying performance gains}
To study performance changes across masking rates, we first choose 60\% as our new standard masking rate. We chose this number since it is the median masking rate threshold, indicating that it is a good compromise between the various tasks. Then, we calculate the change in task scores by increasing the masking rate from 15\% to 60\% and plot the results in figure \ref{img:results_delta}.

We observe that increasing the masking rate improves performance: the change in performance is positive across all configurations except for one. Even the lowest points on the error bars are all positive except for two. These results show with high confidence that increasing the masking rate improves performance over all strategies and tasks.

Next, we analyze the improvement in performance per task.

\paragraph{MLM-based Tasks} We first analyze MLM-based tasks, namely VQA and NLVR. Within VQA, we see a significant gain of 1.6 points for uniform masking, and an average of 1 point across all other strategies. In NLVR, we see a modest increase of around 0.3 points for all strategies except PMI masking. Since both tasks show gains in performance, we can conclude that increasing the masking rate to 60\% improves performance on MLM-based tasks.

\paragraph{ITM-based Tasks} Next, we look at results of Flickr30k retrieval. Surprisingly, we still see performance improvements when increasing the masking rate. For image retrieval, we see an increase of around 3 points across all masking strategies, and for text retrieval, we see an improvement of 2 points in all strategies except noun-verb masking. We see similar improvements for COCO image and text retrieval, with an increase of around 3 points for all strategies except noun-verb masking. Since this task relies solely on ITM, we can conclude that altering the MLM properties leads to better ITM performance, suggesting a deeper link between MLM and ITM. We leave exploration of this connection to future work.

\subsection{Uniform masking performs best at a high masking rate}
In this section, we demonstrate how uniform masking performs well at high masking rate in two ways: through scalability and competitiveness.

\textbf{Uniform masking is scalable}: We call a masking strategy scalable if shows performance gains upon increasing the masking rate. The more scalable a strategy is, the more performance we get from the same change in masking rate. Scalability is important since we want to utilize as much data as we have, especially since paired image-caption data is scarce. Employing a higher masking rate directly leads to predicting more tokens from the same data point, implying higher data utilization.

We demonstrate the scalability of uniform masking by analyzing performance gains per masking strategy in figure \ref{img:results_delta}. Uniform masking displays the highest increase in performance in VQA, Flickr text retrieval and COCO text retrieval. Moreover, in all other tasks, the difference between the highest performance and uniform masking is insignificant. These results indicate that uniform masking scales well across all tasks considered.

\textbf{Uniform masking is competitive}: We quantify the competitiveness of a masking strategy by calculating the empirical probability that it performs better than any other strategy. We compute this quantity for uniform masking in two steps. First, we gather pairs of downstream performance for a given masking rate by taking the cartesian product of uniform masking and non-uniform masking performance. Then, we compute the fraction of pairs where uniform masking performs better than the other strategy.

Concretely, let $X$ be the set of performace across all strategies and let $U$ be the set of performance of uniform masking such that $U\subset X$. Then the empirical probability $\hat P$ is

\[ \hat P = \frac 1 {|U|(|X| - |U|)} \sum_{u\in U}\sum_{x \in X \backslash U} 1\{ x > u \} \]

\begin{figure*}
  \includegraphics[width=\textwidth]{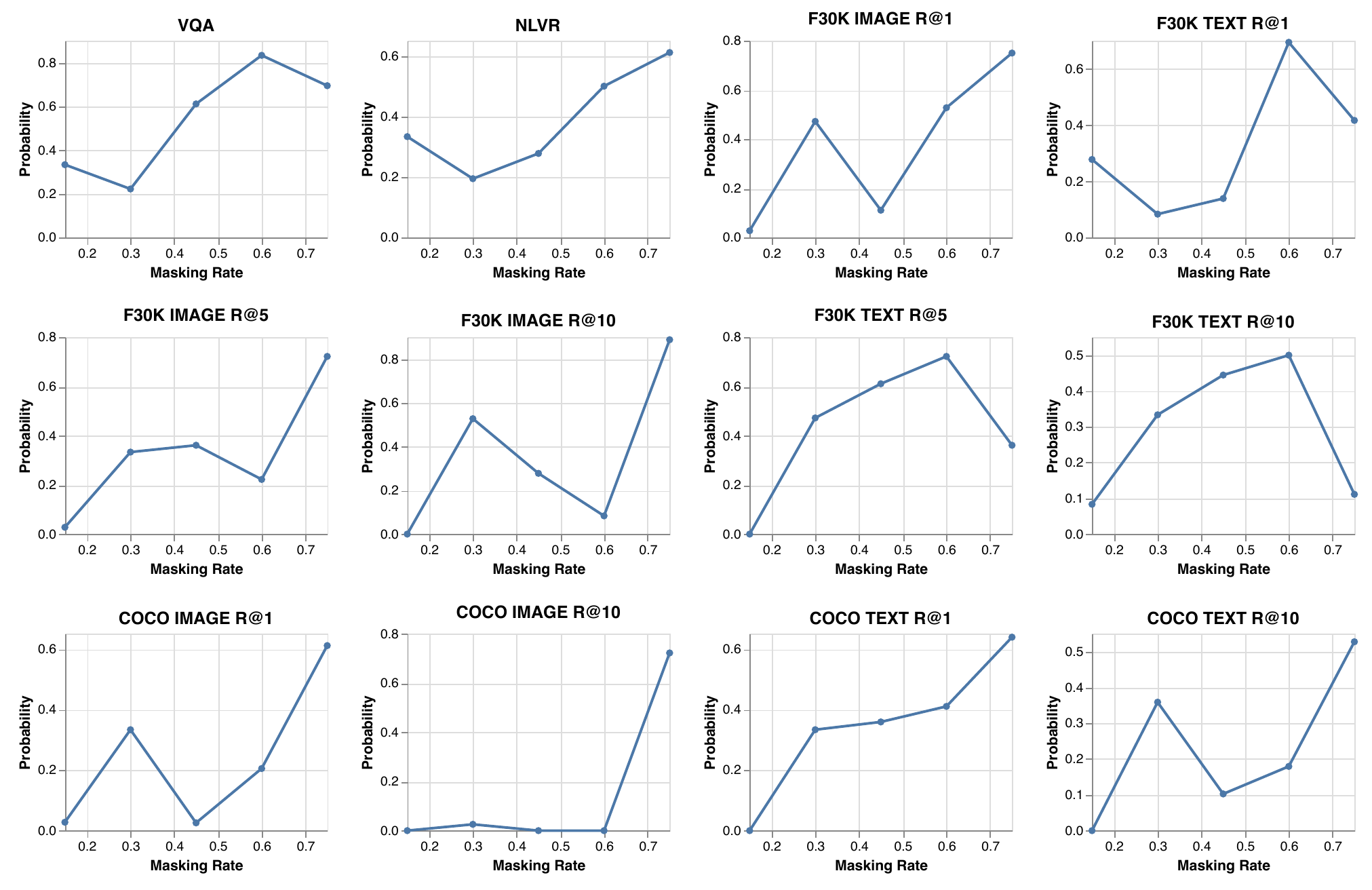}
  \caption{Empirical probability that uniform masking performs better than other masking strategies. Probability increases as we increase masking rate, indicating that uniform masking performs better than other strategies at high masking rates. Extension of table \ref{tbl:uniform}.}
  \label{img:perf}
\end{figure*}

We perform this calculation across all tasks and masking rates and list a summary of the results in table \ref{tbl:uniform}. A more detailed plot with the full results is given in figure \ref{img:perf}.

\begin{table}
\centering
\begin{tabular}{lcc}
  \hline
  Task & 15\% & 60\% \\
  \hline
  VQA & $0.33$ & $\bf0.83$ \\
  NLVR & $0.33$ & $\bf0.50$ \\
  Flickr Image R@1 & $0.03$ & $\bf0.53$ \\
  Flickr Text R@1 & $0.28$ & $\bf0.69$ \\
  COCO Image R@1 & $0.03$ & $\bf0.21$ \\
  COCO Text R@1 & $0.00$ & $\bf0.41$ \\
  \hline
\end{tabular}
\caption{Empirical probability that uniform masking performs better than other masking strategies. Probability increases as we increase masking rate, making uniform masking compare favorably to more complex masking strategies at high masking rates.}
\label{tbl:uniform}
\end{table}

We see that increasing the masking rate makes uniform masking more likely to be the best performer across all masking strategies. Furthermore, most results have a probability greater than $0.5$, indicating that uniform masking is more likely to perform better than other methods at high masking rates. These results conclusively show that increasing the masking rate makes uniform masking perform better than other masking strategies.

\section{Related Work}
Recent work has shown that increasing the masking rate can lead to substantial gains in performance. \citet{he2021mae} discovered conducting self-supervised learning on images requires increasing the masking rate to 80\%. Similarly, \citet{geng2022mmae} trained a multimodal model and found that by increasing the masking rate to a value between 50\% and 90\%, they can improve downstream performance.

Closely related work was performed by \citet{wettig2022should}, who conducted an analysis of masking rate in the context of text-based LM Our work differs from theirs in a few ways. First, we study the masking rate in the multimodal setting, which does not necessarily have the same effect as the text-only setting. We find that we can increase the masking rate much higher than their suggestion of 40\%. Second, we study the effect across multiple masking strategies in depth, something they do not highlight in their paper. Finally, we point out the link between MLM and ITM which is multimodal specific.

\section{Conclusion}
In this paper, we analyzed the effect of masking rate and masking strategy on Vision-Language pretraining. We demonstrated that increasing the masking rate improves performance across masking strategies. In addition, we demonstrated the competitiveness and scalability of uniform masking at high masking rates. From these results, we recommend a masking rate of at least 60\% when training multimodal models. Furthermore, to improve performance, we recommend using uniform masking and increasing the masking rate rather than switching masking strategies.

Excitingly, by increasing the masking rate, we found that MLM has an effect on ITM tasks, indicating deeper ties between the losses. Further work can better quantify this relation and how it changes with model architecture and dataset scale.

\section*{Limitations}
We identify three prominent limitations of our work. Our experiments require a lot of compute resources due to the numerous pretraining and finetuning runs along with a large grid sweep, making it expensive to reproduce our results. Moreover, even with this large sweep, we cannot test all masking strategies. Although we believe that our pick of strategies is quite broad, there is the possibility that new forms of masking strategies will render uniform masking obsolete. Finally, our work can aid the development of more powerful VL understanding models, which could be used to cause societal harm through generation of fake news or harmful content.

\makeatletter
\ifacl@finalcopy
\section*{Acknowledgements}
We thank Yuning Mao for providing feedback on the paper draft.
\fi
\makeatother

\bibliographystyle{acl_natbib}
\bibliography{references}

\clearpage

\section{Appendix}
\begin{table}
\begin{tabular}{ll}
\hline
\bf Parameter & \bf Value \\
\hline
\bf Training Settings \\
\hline
effective\_batch\_size & 4096 \\
per\_gpu\_batch\_size & 64 \\
num\_gpus\_per\_node & 8 \\
num\_nodes & 4 \\
\hline
\bf Image Settings \\
\hline
image transform & pixelbert \\
image\_size & 384 \\
patch\_size & 32 \\
max\_text\_len & 40 \\
tokenizer & bert-base-uncased \\
\hline
\bf Transformer Settings \\
\hline
ViT architecture & vit\_base\_patch32\_384 \\
hidden\_size & 768 \\
num\_heads & 12 \\
num\_layers & 12 \\
mlp\_ratio & 4 \\
drop\_rate & 0.1 \\
\hline
\bf Optimizer Settings \\
\hline
optim\_type & adamw \\
learning\_rate & 1e-4 \\
weight\_decay & 0.01 \\
decay\_power & 1 \\
max\_epoch & 10 \\
warmup\_steps & 2500 \\
lr\_mult & 1 \\
\hline
\end{tabular}
\caption{Hyperparameters for ViLT pretraining}
\label{tbl:hyperparams}
\end{table}

\subsection{Results}
Below we list the resuts from the paper with more metrics and information.

\begin{table*}
\begin{tabular}{lccccc}
\hline
\bf Task & \bf Noun Verb & \bf PMI & \bf Span & \bf Uniform & \bf Whole Word \\
\hline
VQA & $0.55\pm0.03$ & $1.14\pm0.75$ & $0.93\pm0.26$ & $1.64\pm0.46$ & $0.82\pm0.21$ \\
NLVR & $0.31\pm0.48$ & $0.09\pm0.05$ & $0.44\pm0.33$ & $0.41\pm0.30$ & $0.39\pm0.20$ \\
Flickr Image R@1 & $1.11\pm0.52$ & $2.52\pm0.20$ & $2.91\pm0.27$ & $3.73\pm0.03$ & $3.75\pm0.46$ \\
Flickr Image R@5 & $0.05\pm0.46$ & $2.16\pm0.45$ & $1.52\pm0.51$ & $2.57\pm0.54$ & $2.04\pm0.23$ \\
Flickr Image R@10 & $0.22\pm0.44$ & $1.65\pm0.24$ & $1.36\pm0.14$ & $1.59\pm0.27$ & $1.53\pm0.20$ \\
Flickr Text R@1 & $-1.80\pm0.46$ & $2.93\pm0.50$ & $2.90\pm0.70$ & $3.20\pm0.66$ & $1.57\pm1.07$ \\
Flickr Text R@5 & $-0.07\pm0.15$ & $0.73\pm0.75$ & $0.77\pm0.31$ & $1.70\pm0.44$ & $0.90\pm0.17$ \\
Flickr Text R@10 & $-0.23\pm0.23$ & $0.50\pm0.20$ & $0.37\pm0.61$ & $0.83\pm0.47$ & $0.43\pm0.49$ \\
COCO IMAGE R@1 & $0.62\pm0.22$ & $2.98\pm0.10$ & $3.02\pm0.32$ & $3.00\pm0.10$ & $3.16\pm0.31$ \\
COCO IMAGE R@10 & $0.83\pm0.28$ & $2.00\pm0.09$ & $2.39\pm0.32$ & $2.50\pm0.03$ & $2.87\pm0.40$ \\
COCO IMAGE R@5 & $0.71\pm0.29$ & $2.48\pm0.25$ & $3.05\pm0.43$ & $3.17\pm0.23$ & $3.94\pm0.23$ \\
COCO TEXT R@1 & $0.01\pm1.05$ & $3.43\pm0.45$ & $2.85\pm0.79$ & $4.11\pm0.72$ & $3.63\pm0.34$ \\
COCO TEXT R@10 & $0.53\pm0.48$ & $2.01\pm0.14$ & $0.53\pm0.15$ & $2.10\pm0.26$ & $1.81\pm0.54$ \\
COCO TEXT R@5 & $0.05\pm0.53$ & $2.60\pm0.54$ & $1.25\pm0.29$ & $2.58\pm0.57$ & $2.83\pm0.58$ \\
\hline
\end{tabular}
\caption{Change in performance when increasing masking rate from 15\% to 60\% across various tasks and masking strategies. Error represents standard error of the mean. Almost all results are positive, indicating that increasing the masking rate improves performance. This is figure \ref{img:results_delta} in tabular form with more data.}
\end{table*}

\begin{figure*}
  \includegraphics[width=\linewidth]{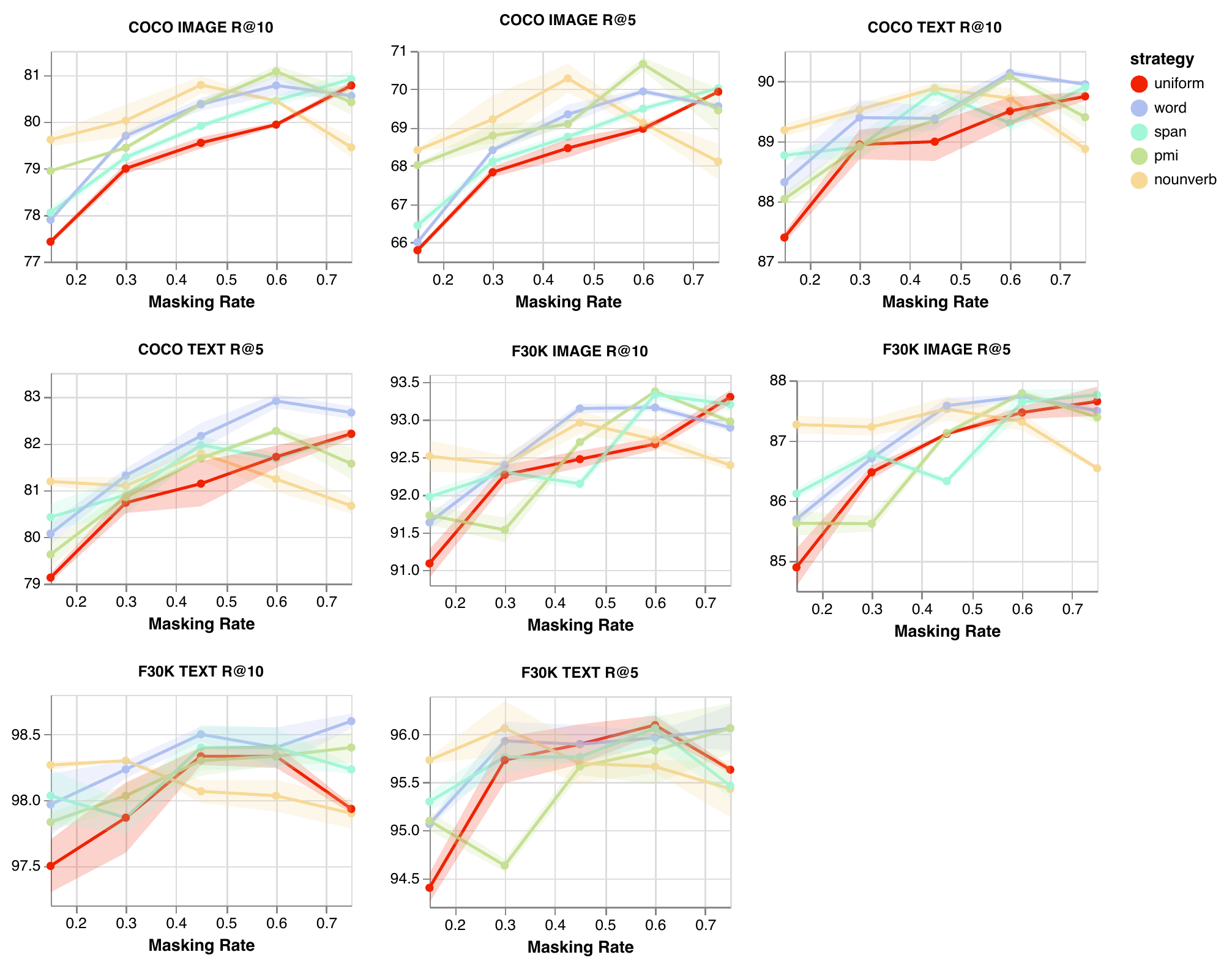}
  \caption{Downstream tasks performance vs masking probability across all strategies. Bars show standard error of the mean. Trend is generally upwards, indicating that increasing masking rate leads to performance gains on a suite of downstream tasks. Continuation of figure \ref{img:results_all}.}
\end{figure*}

\end{document}